\documentclass[letterpaper, 10 pt, conference]
{ieeeconf}  
\usepackage{amsmath}
\usepackage{amssymb}
\usepackage{amsfonts}
\usepackage{graphicx} 
\usepackage{tikz}
\usetikzlibrary{positioning}
\usetikzlibrary{arrows.meta, positioning}
\newcommand{\tightbmatrix}[1]{\begin{bmatrix}#1\end{bmatrix}}

\usepackage{booktabs}
\usepackage[utf8]{inputenc}
\usepackage{textgreek}
\usepackage{subcaption}
\usepackage{graphicx}   
\usepackage{tikz}
\usetikzlibrary{arrows.meta, decorations.pathmorphing}
\usepackage{tikz}
\usepackage{amsmath}
\usetikzlibrary{positioning}
\usepackage{wrapfig}
\usepackage{siunitx}
\usepackage{wrapfig}
\newenvironment{remark}{\noindent\textbf{Remark.}\itshape}{\par}

\IEEEoverridecommandlockouts                              
\title{\LARGE \bf
Learning-Based Modeling of Soft Robots via Cosserat Rod Theory
}

\author{Mohammad Ali, Nithin Senthur Kumar, Eric J. Barth, and Thomas Beckers
\thanks{M. Ali and T. Beckers are with the Department of Computer Science, 
          Vanderbilt University, Nashville, TN 37235, USA,
          {\tt\small mohammad.ali@vanderbilt.edu, thomas.beckers@vanderbilt.edu}}%
\thanks{N. S. Kumar and E. J. Barth are with the Department of Mechanical Engineering, 
          Vanderbilt University, Nashville, TN 37235, USA, 
          {\tt\small nithin.s.kumar@vanderbilt.edu, eric.j.barth@vanderbilt.edu}}%
}

\begin{document}

\maketitle
\thispagestyle{empty}
\pagestyle{empty}

\begin{abstract}

Modeling soft robot dynamics is challenging due to their continuum structure and typically nonlinear dynamics. Creating models based on first-order principles is typically time-demanding, and their expressiveness is limited, whereas data-driven models lack interpretability and physical consistency. This work aims to overcome these challenges by introducing a port--Hamiltonian Gaussian Process Regression framework for learning and simulating the dynamics of planar, rod-like soft robots. In detail, the proposed model integrates Cosserat rod theory and Hamiltonian physics with data-driven inference to preserve the system’s energy structure while accurately learning the rod dynamics. 
Numerical simulations show that we can achieve accurate and energy-consistent representations of a rod-like soft robot, showing the potential for a robust and interpretable pathway for modeling complex continuum mechanics.

\end{abstract}

\section{INTRODUCTION}

Rod-like soft continuum structures such as tentacles, flexible catheters, and continuum manipulators can conform to complex environments and safely interact with humans, making them attractive for applications from minimally invasive surgery to search-and-rescue \cite{calisti2017,cianchetti2018,trivedi2008}. However, their inherent compliance and large deformations make modeling and control significantly more challenging than in rigid robotic systems, as the dynamics are complex and governed by nonlinear partial differential equations (PDEs) \cite{armanini2023}. While high-fidelity finite element simulations can capture complex material behavior, their computational burden limits their use for control. This has motivated reduced-order formulations such as Piecewise Constant Curvature (PCC) and Cosserat rod models, which offer a better balance between physical fidelity and efficiency \cite{armanini2023}. In particular, Cosserat rod theory provides a general and physically consistent description of rod-like soft robots by treating the body as a continuous deformable curve with distributed mass, elasticity, and actuation \cite{dellasantina2023,janabi2021}. Cosserat formulations have been successfully applied to tendon-driven manipulators, concentric-tube robots, and multi-section continuum arms \cite{janabi2021,renda2019}. However, deriving the exact governing equations of Cosserat rod models from first principles is challenging due to the nonlinearities inherent in soft materials.

Alternatively, energy-based models such as port--Hamiltonian (PH) formulations have been extensively used to model and control soft robots \cite{chang2020cdc_ieee,wang2021acc_ieee,caasenbrood2021,ficuciello2010}. The distributed port--Hamiltonian (dPHS) framework generalizes PH theory to PDEs, enabling energy-consistent modeling over spatial domains \cite{macchelli2004} and serving as a foundation for boundary control and energy-based discretization \cite{chun2019}. However, deriving and calibrating the constitutive terms in these models still remains nontrivial. 

In contrast, purely data-driven methods based on neural networks, Koopman operators, and deep recurrent architectures \cite{bruder2019,liu2023iros,qin2024,dellasantina2023} can bypass explicit derivations but typically ignore physical structure, lack stability guarantees, and require large datasets. Recent physics-informed approaches, including PINNs for soft robotic deformation \cite{wang2024pinnray,sun2022,liu2023iros} and equation-embedded networks \cite{deng2024}, improve data efficiency but primarily assume knowledge of the governing equations and lack uncertainty quantification. More recently, Gaussian Process Port--Hamiltonian Systems (GP--PHS) have been proposed to learn unknown energy functions while preserving passivity \cite{beckers2022cdc}, and \cite{tan2024} extended this to PDE-based systems. However, existing GP--PHS approaches have been limited to simple, low-dimensional systems and have not yet been applied to complex dynamics as encountered in soft robotics.
\subsection{Approach and Contribution}
To address these challenges, we propose a hybrid modeling strategy that combines the physical guarantees of the dPHS framework with the flexibility of machine learning. Rather than assuming specific knowledge of the Hamiltonian, we learn the variational derivatives of the Hamiltonian directly from data while maintaining the PH structure of the Cosserat rod model. Specifically, we use Gaussian Process Regression (GPR) to infer these energy-based quantities in a probabilistic, nonparametric, and uncertainty-aware manner, enabling the model to capture nonlinear distributed dynamics without requiring explicit derivation of the governing equations. The main contributions of this paper are twofold. First, we develop the CR-GPR framework that combines PDE-based Cosserat rod (CR) theory with GPR for a physically-consistent learning-based approach for modeling soft robot dynamics with uncertainty quantification. Second, simulations show CR-GPR bridges PDE-based models and data-driven representations, enabling an energy-consistent, data-adaptive framework for modeling of soft robotic dynamics. Its strong prior allows it to excel with small datasets and outperform other data-driven methods.
\section{Problem Formulation}
\label{sec:problem_formulation}
We consider a class of planar soft robots whose diameters are much smaller than their lengths $L_0$ (rod-like) so that their dynamics can be described in continuous space and time by nonlinear partial differential equations (PDEs) of the form
\begin{equation}
  0 = F(z(s,t), u(t)),
  \qquad s \in [0,L_0],~ t \ge 0,
  \label{eq:general_pde}
\end{equation}
where \( z(s,t) \in \mathbb{R}^n \) denotes the distributed state along the arclength (centerline) \( s \) at time \(t\), \( u(t) \in \mathbb{R}^m \) represents external or actuation inputs, and \( F(\cdot) \) is an unknown nonlinear operator encoding the internal elastic, inertial, and dissipative effects of the soft body. The boundary conditions at $s=0$ and $s=L_0$ are assumed to be known and typically reflect clamped, free, or actuated end constraints.

However, the underlying constitutive relations and material parameters such as stiffness, damping, or coupling between deformation modes are not directly measurable, might be nonlinear and unstructured (such as material dependent stress-strain curves), and may vary with configuration. Consequently, the exact functional form of the governing PDE \eqref{eq:general_pde} and their parameters are often only partially known. To overcome this challenge, we assume the existence of spatio-temporal measurements of the robot’s configuration and actuation inputs
\begin{equation}\label{eq:dataset}
\mathcal{D} = \{ z(t_j, s_i),\, u(t_j) \}_{j=1,\ldots,N_t;\, i=1,\ldots,N_s},
\end{equation}
where \( \mathcal{D} \) represents the available dataset sampled from physical experiments or high-fidelity simulations. The dataset can be collected, e.g., by a camera system with a skeletonization algorithm to determine the robot's centerline~\cite{saha2016survey}. Our goal is to identify a physically consistent model of the system’s dynamics based on the dataset $\mathcal{D}$. Specifically, we seek to develop a model which (i) does not require a parametric governing equation, (ii) is physically consistent, (iii) provides a rich structure for future controller design, and (iv) enables uncertainty quantification.
\section{Learning-Based Modeling via Cosserat Rod Theory} 
The central idea is to leverage the collected dataset $\mathcal{D}$ to learn an approximated Cosserat rod model represented in an energy-based formulation via distributed Port--Hamiltonian system theory. In this formulation, the nonlinearities of the governing equations are assembled in the Hamiltonian functional, which inspired us to use GPR to \emph{learn} the Hamiltonian in a probabilistic, nonparametric way. This approach enables physically-consistent learning (in terms of a physical plausible energy evolution) with uncertainty quantification. Next, we  introduce the Cosserat rod theory in distributed Port--Hamiltonian form, followed by the proposed CR-GPR model.
\subsection{Planar Cosserat Rod in Port--Hamiltonian Form}
\begin{figure}[t]
    \centering
    \includegraphics[width=0.45\textwidth]{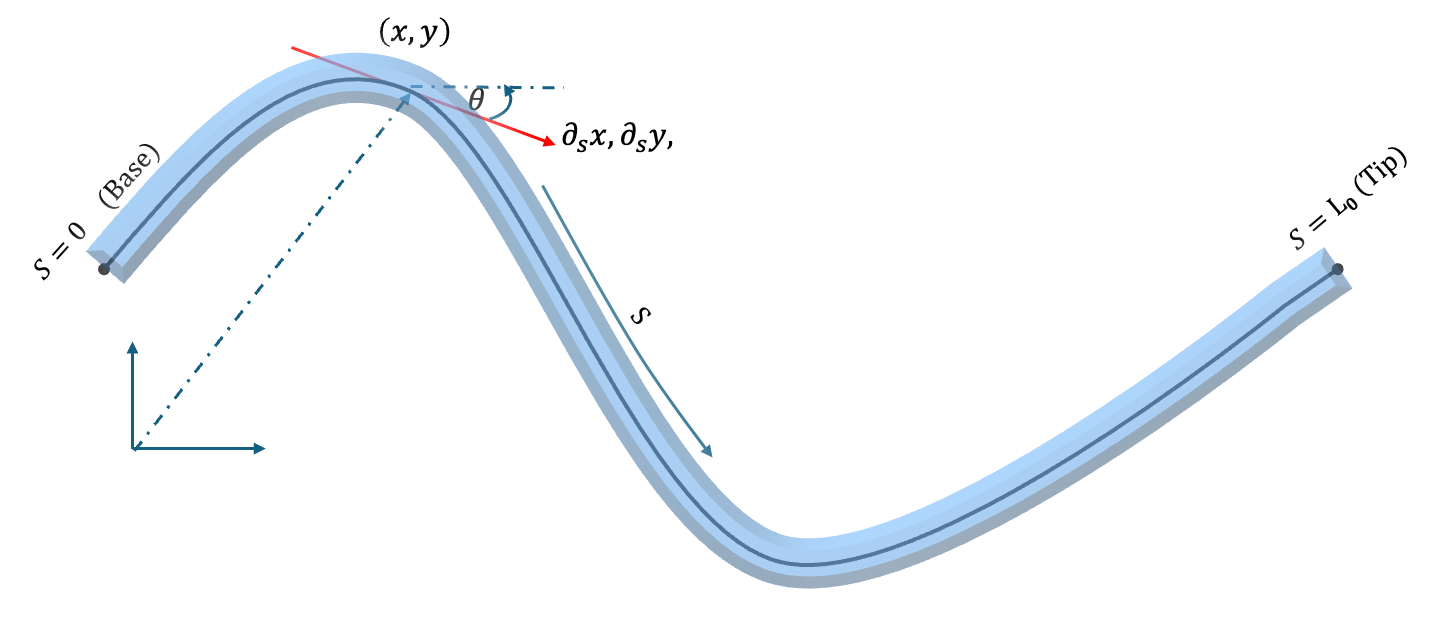}
    \vspace{-0.2cm}
    \caption{Schematic of a planar Cosserat rod.}
    \vspace{-0.5cm}
    \label{fig:cosserat_geometry}
\end{figure}

We consider a planar Cosserat rod of length $L_0$ with arc coordinate $s \in [0,L_0]$ and time $t \geq 0$, see Figure~\ref{fig:cosserat_geometry}. 
The generalized state $z(t,s)\in\mathbb{R}^6$ collects the configuration $q=(x,y,\theta)$ with planar position $(x,y)$ and angle $\theta$ as well as its conjugate momenta $p=(p_x,p_y,p_\theta)$
\begin{equation}
z = [\,x,\;\; y,\;\; \theta,\;\; p_x,\;\; p_y,\;\; p_\theta\,]^{\top},
\end{equation}
where $p_x=\rho A \dot{x},\;\; p_y=\rho A \dot{y},\;\; p_\theta=\rho I \dot{\theta}$ with the material density $\rho$, the cross-sectional area $A$, and the area moment of inertia $I$. Following~\cite{chang2020cdc_ieee}, the total energy of the robot (Hamiltonian functional) is 
the sum of kinetic and stored potential energy $W$ given by
\begin{equation}
\label{eq:H}
H(z)=\tfrac12 \int_{0}^{L_0} \!\left(\tfrac{p_x^2+p_y^2}{\rho A}+\tfrac{p_\theta^2}{\rho I}\right) ds
+\int_{0}^{L_0} W(\nu_1,\nu_2,\kappa)\,ds.
\end{equation}
The stored energy function $W$ is given by
\begin{equation}
W(\nu_1,\!\nu_2,\!\kappa) \!=\! \tfrac{1}{2}\!\big[\!EA(\nu_1\!-\!\nu_1^0)^2 \!+\! GA(\nu_2\!-\!\nu_2^0)^2 \!+\! EI(\kappa\!-\!\kappa^0)^2\big],
\end{equation}
with the strain measures
\begin{equation}
\begin{aligned}
\nu_1 &= \cos\theta\,\tfrac{\partial x}{\partial s}
       + \sin\theta\,\tfrac{\partial y}{\partial s}, \\
\nu_2 &= -\sin\theta\,\tfrac{\partial x}{\partial s}
       + \cos\theta\,\tfrac{\partial y}{\partial s},\quad
\kappa = \tfrac{\partial \theta}{\partial s}.
\end{aligned}
\end{equation}



and material constants $E$ (Young’s modulus) and $G$ (shear modulus). The intrinsic strains are typically $\nu_1^0=1,\;\nu_2^0=0,\;\kappa^0=0$. Then, the rod's dynamics can be written in distributed Port--Hamiltonian form via
\begin{equation}
    \frac{\partial z}{\partial t} = (J - R)\frac{\delta H}{\delta z} + G(z)u,
    \label{eq:dphs_form}
\end{equation}
where $J=-J^\top\in\mathbb{R}^{6\times 6}$ is a skew-symmetric matrix that encodes the interconnection structure, $R=R^\top\in\mathbb{R}^{6\times 6}$ with $R\succeq  0$ is the dissipation operator, and $u$ denotes external inputs with input vector $G(z)\in\mathbb{R}^{6}$. The operator $\delta$ denotes the variational derivative as defined in~\cite{rashad2020}. For the planar rod, the canonical structure matrices are

\begin{equation}\label{eq:JR}
J=\begin{bmatrix}
0_{3} & I_{3}\\
-I_{3} & 0_{3}
\end{bmatrix},
\qquad
R=\begin{bmatrix}
0_{3} & 0_{3}\\
0_{3} & \zeta\,I_{3}
\end{bmatrix}
\end{equation}
with damping coefficient $\zeta\in\mathbb{R}_{>0}$, identity matrix $I_n$ and zero matrix $0_n$ of dimensions $n\times n$. Applying the variational derivative to the total energy~\eqref{eq:H}, the dynamics of the Cosserat rod can be written in dPHS form~\eqref{eq:dphs_form} by
\begin{equation}\label{eq:ph-compact}
\raisetag{-1.0ex}
\makebox[\columnwidth][l]{
\resizebox{\dimexpr\columnwidth-1.6em\relax}{!}{$
\frac{\partial}{\partial t}
\!\underbrace{\tightbmatrix{x\\y\\\theta\\p_x\\p_y\\p_\theta}}_z
\!\!=\!\!
\underbrace{\tightbmatrix{
 0&0&0&1&0&0\\
 0&0&0&0&1&0\\
 0&0&0&0&0&1\\
-1&0&0&-\zeta&0&0\\
 0&-1&0&0&-\zeta&0\\
 0&0&-1&0&0&-\zeta}}_{J-R}
\underbrace{\tightbmatrix{
 -\partial_s(n_x\cos\theta-n_y\sin\theta)\\
 -\partial_s(n_x\sin\theta+n_y\cos\theta)+ \rho A g\\
 -\partial_s m + (\nu_1 n_y-\nu_2 n_x)\\
 p_x/(\rho A)\\
 p_y/(\rho A)\\
 p_\theta/(\rho I)}}_{\frac{\delta H}{\delta z}}
\!\!+\!\!\underbrace{\tightbmatrix{0\\0\\0\\\\u\\\phantom{a}}}_{G(z)u}
$}
}
\end{equation}
where $\partial_s=\frac{\partial}{\partial s}$ and the internal forces and bending moments
\begin{equation}
n_x=EA(\nu_1-\nu_1^0), \;
n_y=GA(\nu_2-\nu_2^0), \;
m=EI(\kappa-\kappa^0).
\end{equation}
 Each element of $\frac{\delta H}{\delta z}$ corresponds to a physical quantity derived from the Hamiltonian functional $H(z)$: $\frac{\delta \hat{H}}{\delta x}$ and $\frac{\delta \hat{H}}{\delta y}$ relate to internal axial and shear forces, $\frac{\delta \hat{H}}{\delta \theta}$ corresponds to the bending moment, and $\frac{\delta \hat{H}}{\delta p_x}$, $\frac{\delta \hat{H}}{\delta p_y}$, $\frac{\delta \hat{H}}{\delta p_\theta}$ capture the linear and angular momentum densities. 
We note that the rod's dynamics written in dPHS form~\eqref{eq:ph-compact} are characterized by relatively simple $J$ and $R$ matrices and all parameters/nonlinearities are part of the variational derivative $\frac{\delta H}{\delta z}$.
\subsection{CR-GPR model}\label{sec:alg}
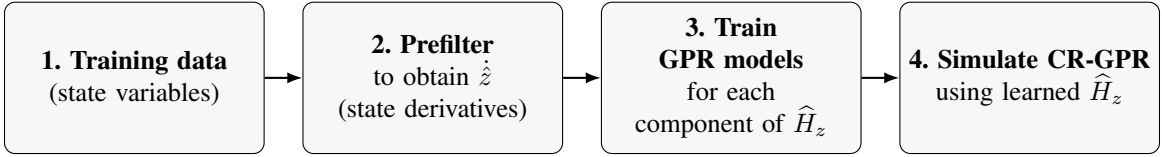
\begin{figure*}[t]
\centering
\begin{tikzpicture}[
  node distance=0.3cm and 0.5cm,
  box/.style={
    draw, rounded corners, align=center,
    minimum height=2.0cm,
    fill=gray!6,
    text width=3.2cm
  },
  arrow/.style={-{Latex[length=2mm]}, thick}
]

\node[box] (n2) {\textbf{1. Training data}\\(state variables)};
\node[box, right=of n2] (n3) {\textbf{2. Prefilter}\\to obtain $\dot{\hat{z}}$\\(state derivatives)};
\draw[arrow] (n2) -- (n3);

\node[box, right=of n3] (n4) {\textbf{3. Train GPR models}\\for each component of $\widehat{H}_z$};
\node[box, right=of n4] (n5) {\textbf{4. Simulate CR-GPR}\\using learned $\widehat{H}_z$};

\draw[arrow] (n3) -- (n4);
\draw[arrow] (n4) -- (n5);

\end{tikzpicture}
\caption{Workflow for learning the variational derivatives of the Cosserat rod dPHS via GPR.}
\label{fig:pipeline}
\end{figure*}

In this section, we introduce our CR-GPR model with the aim to \emph{learn} the variational derivative $\frac{\delta H}{\delta z}$ based on the dataset $\mathcal{D}$ as defined in~\eqref{eq:dataset}. Figure~\ref{fig:pipeline} illustrates the overall workflow.
For notational simplicity, we denote the variational derivative of the Hamiltonian as
$\frac{\delta H(z)}{\delta z} \triangleq H_z$ and its estimate by $\frac{\delta \widehat H(z)}{\delta z} \triangleq \widehat{H}_z$. In the following, we discuss each step of the workflow.









\paragraph{1. Training data} We consider a spatio–temporal dataset consisting of distributed state variables and corresponding actuation inputs,
\begin{equation}
\mathcal{D} = \{ z(t_j, s_i),\, u(t_j) \}_{j=1,\ldots,N_t;\, i=1,\ldots,N_s}.
\end{equation}
The dataset spans $N_t$ time steps and $N_s$ spatial points, corresponding to the sequence of actuation inputs $\{u(t_1), u(t_2), \ldots, u(t_{N_t})\}$ applied over time, where each $z(t_j, s_i)$ contains the state components 
$[x,\, y,\, \theta,\, p_x,\, p_y,\, p_\theta]^\top$ observed at discrete spatial points $s_i$ and time instants $t_j$.
\begin{remark}
    In practice, we can typically observe only the position $[x(s,t),y(s,t)]$ at some spatial and temporal points. However, the angle $\theta$ can be numerically approximated by trigonometry based on positions at two different spatial points and $p_x,\, p_y,\, p_\theta$ can be estimated if the robot's mass distribution is known.
\end{remark}

\paragraph{2. Prefilter} To learn the variational derivative $H_z$ in~\eqref{eq:ph-compact}, we require knowledge of the time derivatives of the state $\dot{z}(t,s)$. In general, we can employ any filtering algorithm for this purpose, but here we focus on a simple method based on GPR. For each component $z_k,k=1,\ldots 6$ of the state $z$, an independent GPR model is trained to learn the mapping $(t, s) \mapsto z_i(t, s)$ based on the dataset $\mathcal{D}$. The GPR model uses a differentiable kernel and provides continuous, differentiable approximations of the spatio–temporal trajectories of each state component. This allows the approximation of time and spatial derivatives $\widehat{\partial_t z}(t,s),\widehat{\partial_s z}(t,s)\in\mathbb{R}^6$ by differentiating the GP-predicted trajectories with respect to time (which can be done analytically). In particular, we create the following dataset
\begin{equation}
\begin{aligned}
\mathcal{E}
= \Big\{\, &
\hat{z}(t_j,s_i),\;
\widehat{\partial_t z}(t_j,s_i),\;
\widehat{\partial_t^2 z}(t_j,s_i),\\
&
\widehat{\partial_s x}(t_j,s_i),\;
\widehat{\partial_s y}(t_j,s_i),\;
\widehat{\partial_s^2 \theta}(t_j,s_i)
\,\Big\},\\
& \qquad j=1,\ldots,N_t',\;\; i=1,\ldots,N_s'.
\end{aligned}
\label{eq:derivative_dataset}
\end{equation}

which will be used in the next step of the pipeline to learn the elements in the variational derivative $H_z$. Here, the operator $\hat{\cdot}$ refers the posterior mean of GPR based on dataset $\mathcal{D}$ and $\partial_s,\partial_{s^2}$ its first and second derivative with respect to the spatial $s$ at $N_s'$ spatial and $N_t'$ temporal points. In this way, we can up- or downsample the number of training points from $\mathcal{D}$ to~\eqref{eq:derivative_dataset} if we choose $N_t\neq N'_t$ or $N_s\neq N'_s$. Eventually, the dataset $\mathcal{E}$ encapsulates the spatio–temporal evolution of the distributed system and serves as the training basis for reconstructing the variational derivatives. 

\paragraph{3. Train GPR models} The variational derivative $\widehat{H}_z$ can be estimated by rewriting the Cosserat rod dPHS~\eqref{eq:dphs_form} as
\begin{equation}
\widehat{H}_z = (J - R)^{-1} [ \widehat{\partial_t z}(t_j, s_i) - G(\hat{z}(t_j, s_i))u(t_j)] ,
\label{eq:estimated_variational_derivative}
\end{equation}
where $J-R$ is invertible by definition. We employ GPR models to approximate each element of the estimated variational derivative vector $\widehat{H}_z$ based on the dataset $\mathcal{E}$. For the first two components, $\frac{\delta \hat{H}}{\delta x}$ and $\frac{\delta \hat{H}}{\delta y}$, each GPR model learns the mapping between local deformation features and the corresponding internal force distributions. Specifically, we learn the mapping
\begin{equation}
\text{GPR }1,\!2\!:\!
\big[\hat{\theta}(t_j,\!s_i),\,
\widehat{\partial_s x}(t_j,\!s_i),\,
\widehat{\partial_s y}(t_j,\!s_i)\big]
\!\mapsto\! \widehat{H}_{z,k}, k\!=\!1,\!2.
\end{equation}

that maps the local orientation, stretch and shear deformation along the rod to the distributed internal forces,
$\frac{\delta \hat{H}}{\delta x}$ and $\frac{\delta \hat{H}}{\delta y}$, respectively. The third GPR corresponds to the variational derivative
$\frac{\delta \hat{H}}{\delta \theta}$ and maps from the local orientation, neighboring segment angles (providing spatial coupling), curvature, and strain features
\begin{equation}
\begin{aligned}
\text{GPR }3:\;
\big[
\hat{\theta}(t_j,s_i),\;
\hat{\theta}(t_j,s_{i-1}),\;
\hat{\theta}(t_j,s_{i+1}),\\
\qquad\widehat{\partial_{s^2}\theta}(t_j,s_i),\;
\widehat{\partial_s x}(t_j,s_i),\;
\widehat{\partial_s y}(t_j,s_i)
\big]
\mapsto \widehat{H}_{z,3}.
\end{aligned}
\end{equation}
This model thus learns the spatial coupling between curvature variations and internal moment 
distribution within the dPHS structure. The remaining three GPR models map
\begin{equation}
\begin{aligned}
    \text{GPR 4-6: }\hat{p}_x(t_j, s_i)\mapsto \widehat{H}_{z,4},\,\, \hat{p}_y(t_j, s_i)\mapsto \widehat{H}_{z,5},\,\,\\
    \hat{p}_\theta(t_j, s_i)\mapsto \widehat{H}_{z,6}.
\end{aligned}
\end{equation}
For each GPR, we employ the squared exponential kernel, as the Hamiltonian and its derivatives are expected to be smooth, and optimize the hyperparameters via marginal log-likelihood maximization~\cite{williams1995}. These learned mappings approximate the relationships between the distributed momenta and
their conjugate quantities, thereby completing the set of GPR models that collectively
represent the variational derivatives $\widehat{H}_{z}$.

\begin{remark}
We choose GPs over alternative function approximators (e.g., neural networks) for three reasons: (i)~the SE kernel encodes a smoothness prior that matches the expected regularity of Hamiltonian gradients; (ii)~GPs provide closed-form posterior predictions without iterative optimization at inference time, enabling direct substitution into the PDE solver; and (iii)~the GP posterior naturally yields calibrated uncertainty estimates that can be propagated to trajectory-level confidence intervals via RFF sampling (see below), which is central to the UQ capability of CR-GPR.
\end{remark}

\paragraph{4. Simulate CR-GPR} Finally, the learned GPR models for $\widehat{{H}}_z$ collectively approximate the variational derivatives of the Cosserat rod, providing a data-driven representation that can be directly employed for reconstructing the rod’s distributed dynamics within the dPHS framework. For this purpose, we need to solve the learned nonlinear PDE
\begin{equation}\label{eq:nonpde}
    \partial_t z = (J - R)\widehat{{H}}_z + G(z)u
\end{equation}
with its boundary and initial conditions, which can be done by standard techniques such as discretization along the 1d spatial domain $s\in[0,L_0]$ and then solve the set of nonlinear ordinary differential equations.

\paragraph{Uncertainty Quantification via RFF Sampling.}
A key advantage of the GP-based framework is that it provides principled uncertainty quantification.
However, pointwise GP posterior variance is insufficient here: the learned functions $\widehat{H}_z$ enter a nonlinear ODE, and the uncertainty of interest is over entire \emph{trajectories}, not individual function evaluations.
To obtain trajectory-level uncertainty, we approximate GP samples of $\widehat{{H}}_z$ using Random Fourier Features (RFF)~\cite{wilson2020}.
For a GP with squared exponential kernel, the RFF approximation is
$f(x) \approx \phi(x)^\top w$, where $\phi(x) = \sqrt{{2\sigma_f^2}/{D}} \cos(Wx + b)$,
$W \sim \mathcal{N}(0, \mathrm{diag}(1/\ell^2))$, and $b \sim \mathrm{Uniform}[0, 2\pi]$.
Given training data, the posterior over weights is $w \sim \mathcal{N}(\mu_w, \Sigma_w)$.
Sampling $S$ weight vectors yields $S$ \emph{joint} function realizations that are correlated across input space, a property that pointwise variance alone cannot provide.
Each sampled variational derivative $\bar{H}_z^{(s)}$ is substituted into~\eqref{eq:nonpde}, and the resulting PDE is solved to produce $S$ trajectory predictions.
The ensemble of trajectories characterizes the predictive distribution, from which confidence intervals and coverage statistics can be estimated.

\section{Simulation}
\label{sec:implementation}

\begin{table}[!t]
\vspace{0.2cm}
\caption{Physical and Simulation Parameters}
\label{tab:sim_params}
\centering
\setlength{\tabcolsep}{6pt}
\renewcommand{\arraystretch}{1.5}
\begin{tabular}{lcc}
\toprule
\textbf{Parameter} & \textbf{Symbol / Unit} & \textbf{Value} \\
\midrule
Rod length & $L_0$ [m] & 0.5 \\
Number of segments & $N_s$ & 25 \\
Spatial step & $\Delta s = L_0/N_s$ [m] & 0.02 \\
Cross-section radius & $r$ [m] & 0.01 \\
Cross-section area & $A = \frac{\pi r^2}{4}$ [m$^2$] & --- \\
Moment of inertia & $I = A^2/(4\pi)$ [m$^4$] & --- \\
Young’s modulus & $E$ [Pa] & $3\times10^6$ \\
Shear modulus & $G = E/(2(1+\nu))$ [Pa] & --- \\
Density & $\rho$ [kg/m$^3$] & 1000 \\
Damping coefficient & $\zeta$ & $10^{-3}$ \\
Gravity & $g$ [m/s$^2$] & 9.81 \\
\bottomrule
\end{tabular}
\end{table}

In this section, we validate the presented CR-GPR method on a numerical example. All experiments were conducted in MATLAB~R2024b. As ground truth, a planar Cosserat rod  with gravity and damping is simulated, governed by the dynamics in~\eqref{eq:ph-compact}. The physical and material parameters are summarized in Table~\ref{tab:sim_params}.
The system was initialized as a straight rod inclined at~$85^{\circ}$ with a clamped base and a free tip boundary condition. The dynamics are solved using a Störmer--Verlet integrator with a time step of $\Delta t = 10^{-4}\,\text{s}$, which preserves the symplectic structure of the Hamiltonian system. The gravity acts as constant input $u$ to the system.
\paragraph{Step 1 of Section~\ref{sec:alg}} For the dataset $\mathcal{D}$, we record a single trajectory, containing 4001 time samples of the full state vector $z$ over a simulation time of $\SI{2}{\second}$. The rod was discretized into \(N = 25\) segments over \(L_0 = \SI{0.5}{\meter}\), resulting in a spatial resolution of 
\(\Delta s = L_0/N = \SI{0.02}{\meter}\). The boundary conditions are set to clamped at the base and free at the tip such that $x(0,t)=0, y(0,t)=0,\theta(0,t)=85^{\circ}, n_x(L_0,t)=0,n_y(L_0,t)=0, m(L_0,t)=0 $.

\paragraph{Step 2 of Section~\ref{sec:alg}} Six GPR models were trained to learn the state evolution over time and its spatial and temporal derivatives to create the dataset $\mathcal{E}$ as in~\eqref{eq:derivative_dataset}. All models employed a Matérn~5/2 
kernel with a pure quadratic basis function. The Matérn~5/2 kernel was chosen because it is twice mean-square differentiable, ensuring that the GP posterior and its analytically computed time and spatial derivatives are continuous~\cite{williams1995}. All models were trained with standardized inputs and a lower noise bound of~$10^{-9}$. Subsequently, the posterior means of the GPR outputs were differentiated in time and space.

\paragraph{Step 3 of Section~\ref{sec:alg}} Six GPR models with squared exponential kernels are trained to learn each component of the variational derivative~${H}_z$. The hyperparameters (kernel length scales~$\ell$, signal variance~$\sigma_f^2$, and noise variance~$\sigma_n^2$) are optimized by maximizing the marginal log-likelihood~\cite{williams1995}. Because the training targets are derived via finite-difference approximation on a spatial grid with step~$\Delta s = L_0/N$, they carry structured discretization errors of order~$\mathcal{O}(\Delta s^2)$. We therefore impose a lower bound on~$\sigma_n$ consistent with the estimated truncation error ($\sigma_n \geq 1.8\!\times\!10^{-3}$ for $\delta H/\delta\theta$; $\sigma_n \geq 3.4\!\times\!10^{-2}$ for $q_x,q_y$) to prevent the GP from overfitting to these numerical artifacts, following standard practice~\cite{williams1995}. Table~\ref{tab:hyperparameters} summarizes all learned hyperparameters.

\begin{table*}[t]
\centering
\caption{GPR hyperparameters for the CR-GPR pipeline. All hyperparameters are optimized by maximizing the marginal log-likelihood. State-variable GPs use Mat\'{e}rn-5/2 kernels for time-series smoothing; gradient models use squared exponential (SE) kernels.}
\label{tab:hyperparameters}
\footnotesize
\setlength{\tabcolsep}{5pt}
\renewcommand{\arraystretch}{1.3}
\begin{tabular}{lcccccc}
\toprule
\textbf{Model} & \textbf{Kernel} & \textbf{Basis} & $\boldsymbol{\sigma_f}$ & $\boldsymbol{\ell}$ & $\boldsymbol{\sigma_n}$ & $\boldsymbol{n}$ \\
\midrule
\multicolumn{7}{l}{\textit{Gradient models (learned from state data)}} \\
$q_x$ & ARD-SE & PureQuadratic & 2.28 & [3.22, 2.47, 181] & 3.4e-02 & 7910 \\
$q_y$ & ARD-SE & PureQuadratic & 1.66 & [3.11, 1.84, 131] & 3.4e-02 & 8652 \\
$\delta H/\delta\theta$ & ARD-SE & Constant & 1.14 & [0.97, 1.86, 21.7, 4.28, 2.04, 51.1] & 1.8e-03 & 18975 \\
$\delta H/\delta p_x$ & SE & None & 134.7 & 61.8 & 9.3e-03 & 966 \\
$\delta H/\delta p_y$ & SE & None & 698.3 & 391.1 & 1.1e-02 & 966 \\
$\delta H/\delta p_\theta$ & SE & None & 8.05 & 54.5 & 1.6e-01 & 980 \\
\midrule
\multicolumn{7}{l}{\textit{State-variable GPs (time-series smoothing, 1D input $t$)}} \\
$x(t),\, y(t),\, \theta(t)$ & Mat\'{e}rn-5/2 & Quadratic & -- & opt. & $\geq 10^{-9}$ & 800/stride \\
$p_x(t),\, p_y(t),\, p_\theta(t)$ & Mat\'{e}rn-5/2 & Quadratic & -- & opt. & $\geq 10^{-9}$ & 1250--2500 \\
\bottomrule
\end{tabular}
\end{table*}

\paragraph{Step 4 of Section~\ref{sec:alg}:} The posterior means of the learned GPR models were integrated into the PHS structure to reconstruct the rod dynamics as in~(\ref{eq:nonpde}). As a test case, we evaluate the learned model on six unseen initial angles ($\theta_0 \in \{60^\circ, 125^\circ, 160^\circ, 180^\circ, 225^\circ, 325^\circ\}$), none of which overlap with the training angle~($85^\circ$), with the rod starting straight at each angle.. Hence, the model aims to \emph{generalize} across initial conditions that span $265^\circ$ of angular range. The CR-GPR simulation uses the same Störmer--Verlet integrator and time step ($\Delta t = 10^{-4}\,\text{s}$) as the ground truth model.

\section{Results and Discussion}
Table~\ref{tab:multiangle_accuracy} reports $\text{RMSE}(\theta)$ and~$R^2$ for each test angle. We focus on~$\theta$ because segment orientations fully determine the rod shape via integration of $\cos\theta$ and $\sin\theta$ from the clamped base, making it the most informative single-variable metric.

\begin{table}[b]
\caption{CR-GPR accuracy across six unseen test angles. The model is trained on a single trajectory ($\theta_0=85^\circ$) and evaluated on each angle independently.}
\label{tab:multiangle_accuracy}
\centering
\setlength{\tabcolsep}{6pt}
\renewcommand{\arraystretch}{1.4}
\begin{tabular}{lcccccc|c}
\toprule
& $60^\circ$ & $125^\circ$ & $160^\circ$ & $180^\circ$ & $225^\circ$ & $325^\circ$ & \textbf{Mean} \\
\midrule
RMSE($\theta$) & 0.316 & 0.244 & 0.224 & 0.210 & 0.144 & 0.193 & \textbf{0.222} \\
$R^2$          & 0.924 & 0.951 & 0.934 & 0.928 & 0.900 & 0.876 & \textbf{0.919} \\
\bottomrule
\end{tabular}
\end{table}

\begin{figure}[h!]
    \centering
    \begin{minipage}{0.5\columnwidth}
        \centering
        \includegraphics[width=\linewidth]{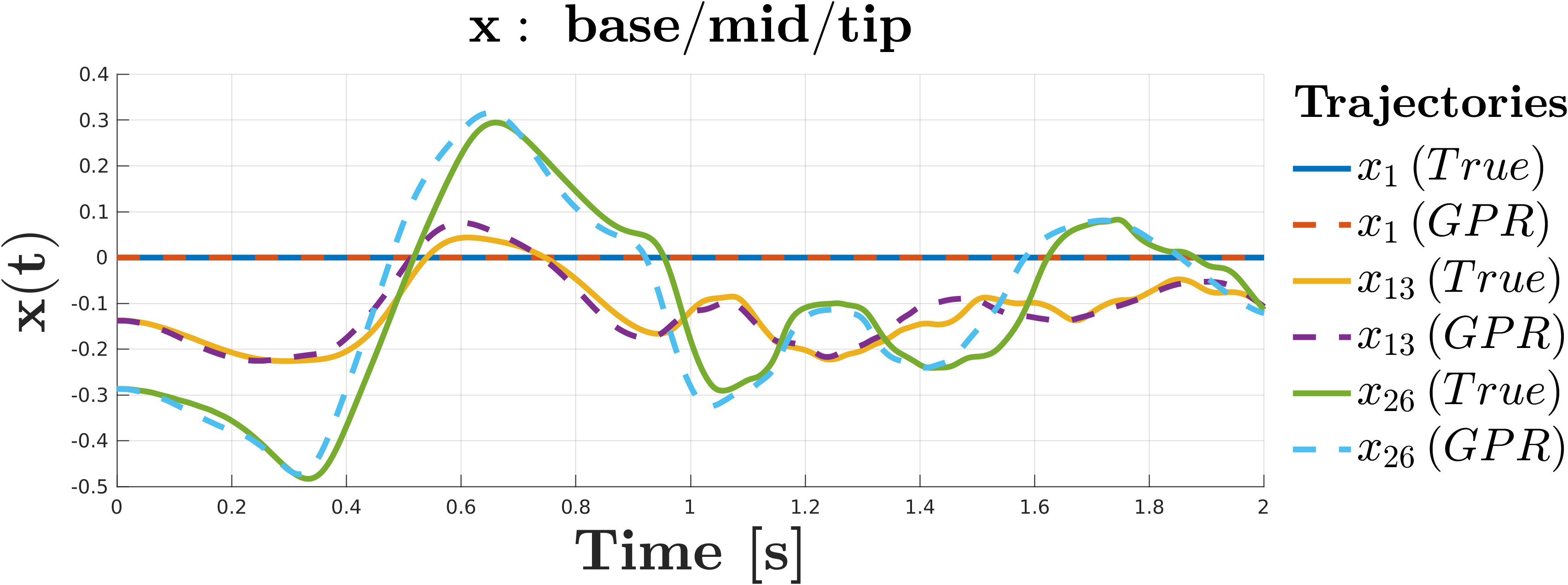}\\[-1.3pt]
        \includegraphics[width=\linewidth]{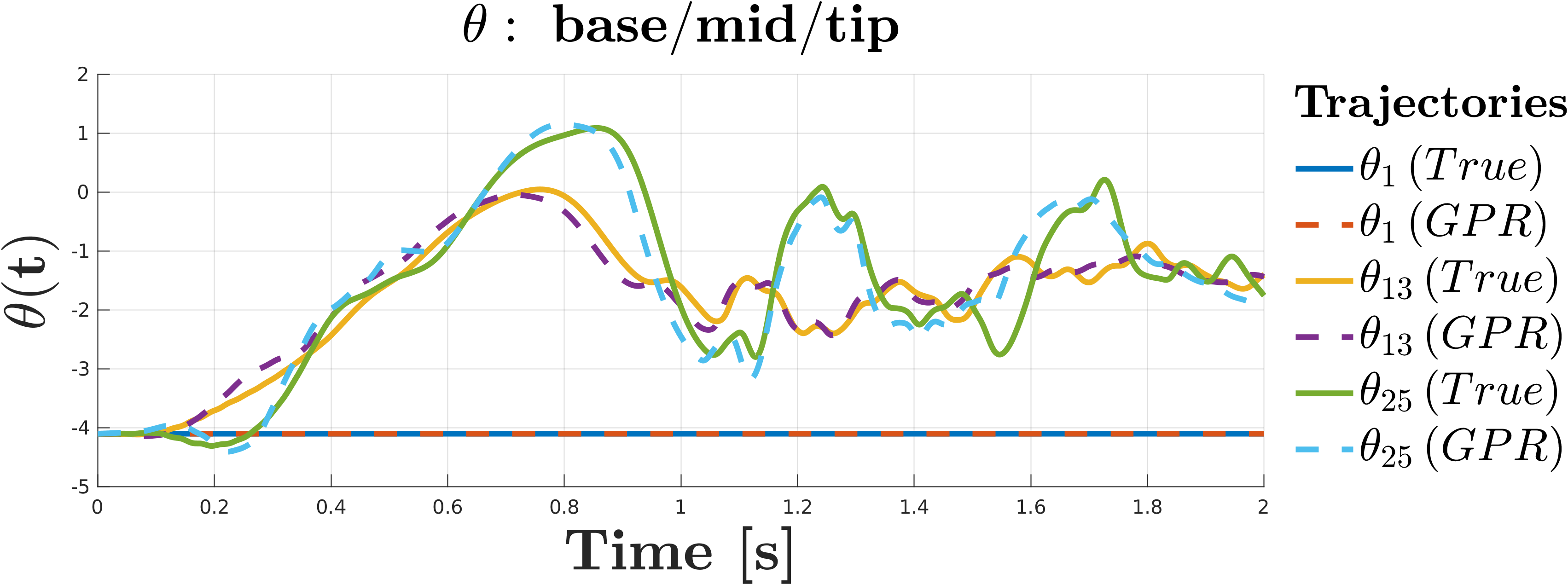}\\[-1.3pt]
        \includegraphics[width=\linewidth]{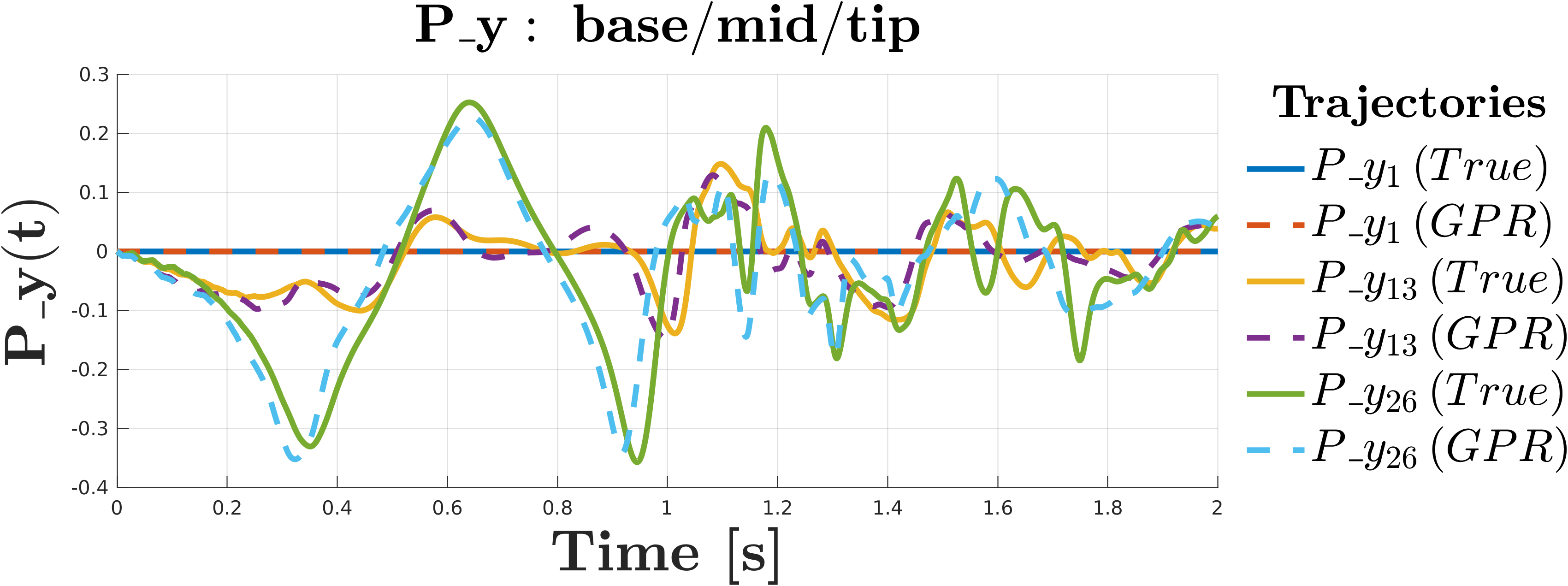}
    \end{minipage}\hfill
    \begin{minipage}{0.5\columnwidth}
        \centering
        \includegraphics[width=\linewidth]{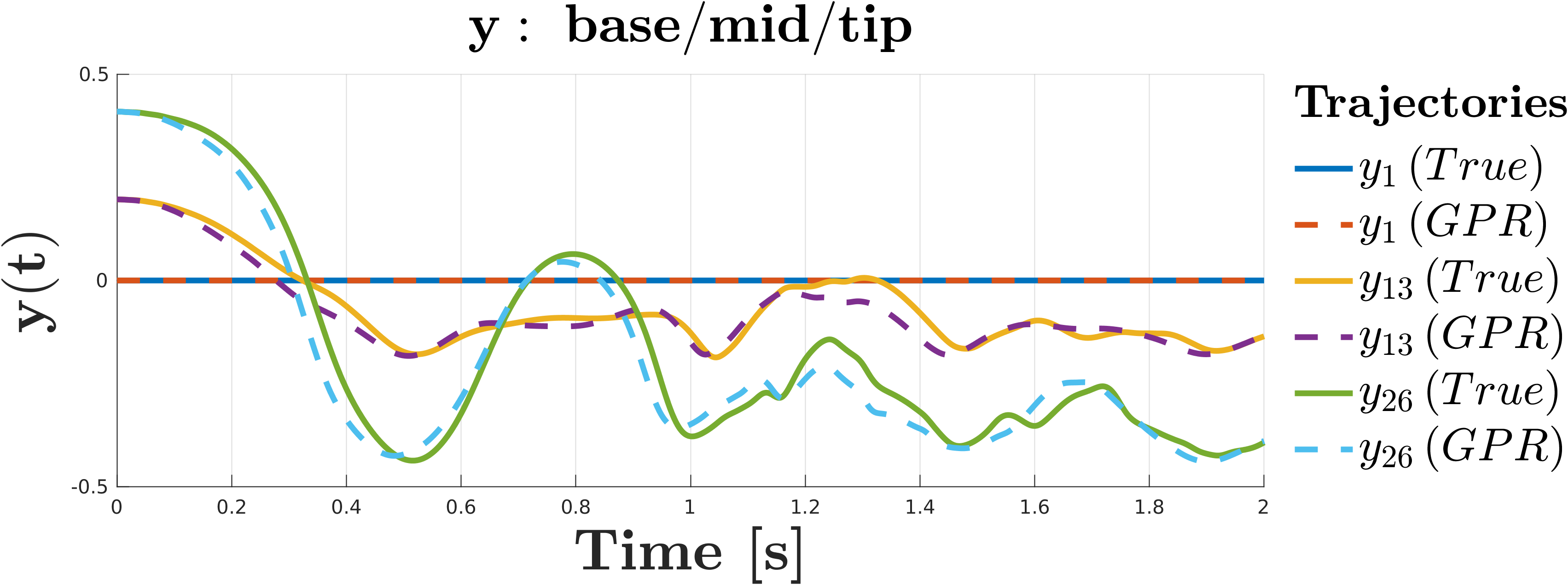}\\[-1.3pt]
        \includegraphics[width=\linewidth]{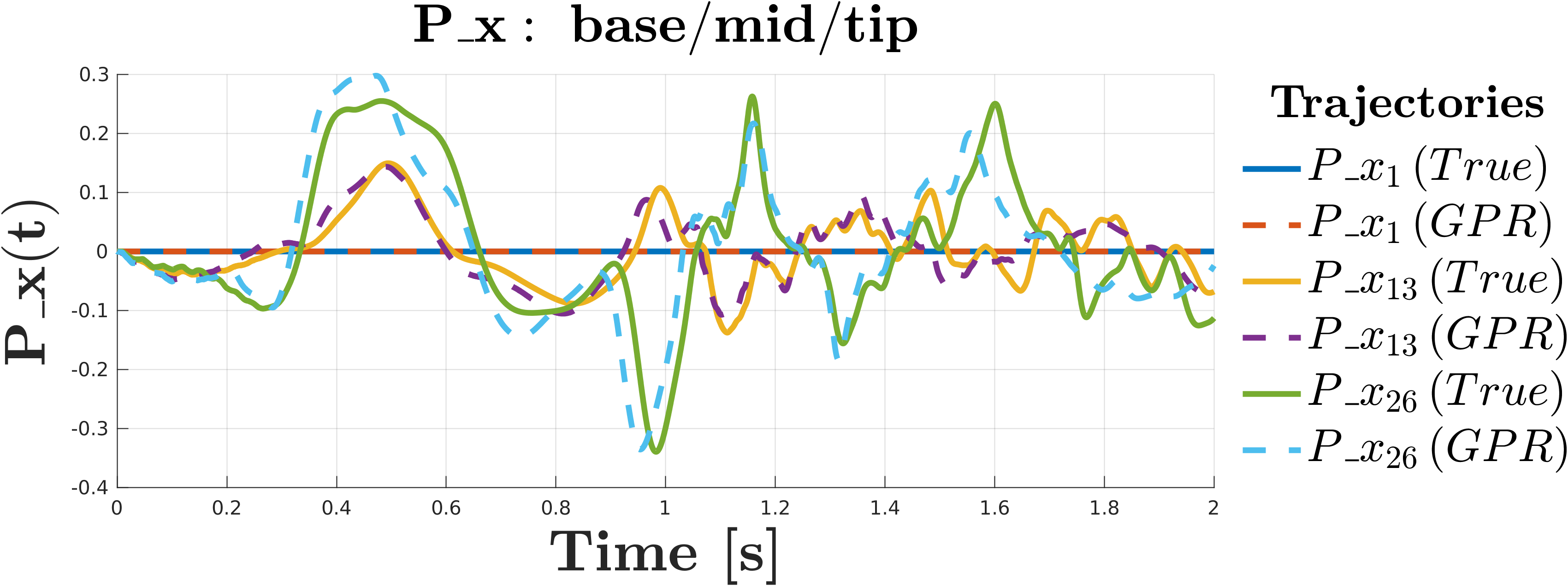}\\[-1.3pt]
        \includegraphics[width=\linewidth]{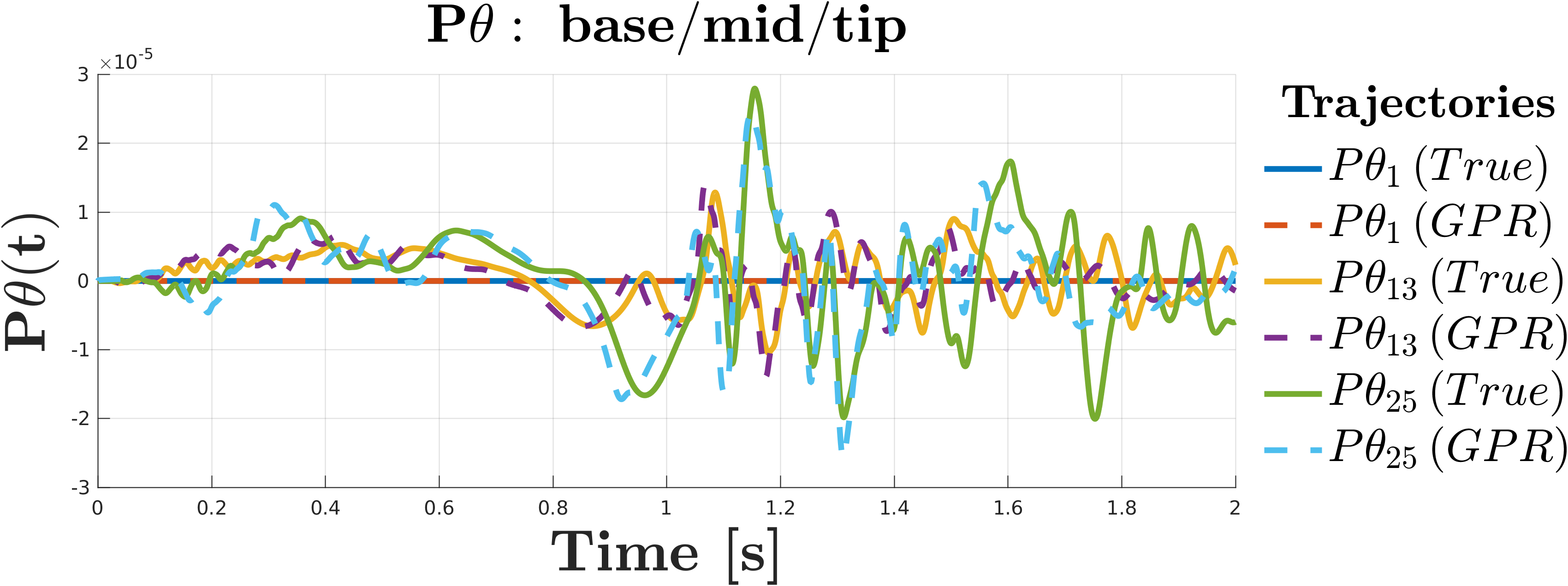}
    \end{minipage}
    \caption{Ground truth vs. CR-GPR simulation overlays at base, mid, and tip nodes for all state variables ($\theta_0 = 125^\circ$).
    Solid lines: ground truth; dashed lines: CR-GPR predictions.}
    \label{fig:overlays_all}
\end{figure}

\begin{figure*}[t!]
    \centering

    \begin{minipage}[t]{0.5\textwidth}
        \hspace{-0.5cm}\vspace{-0.2cm}\includegraphics[width=1.15\textwidth]{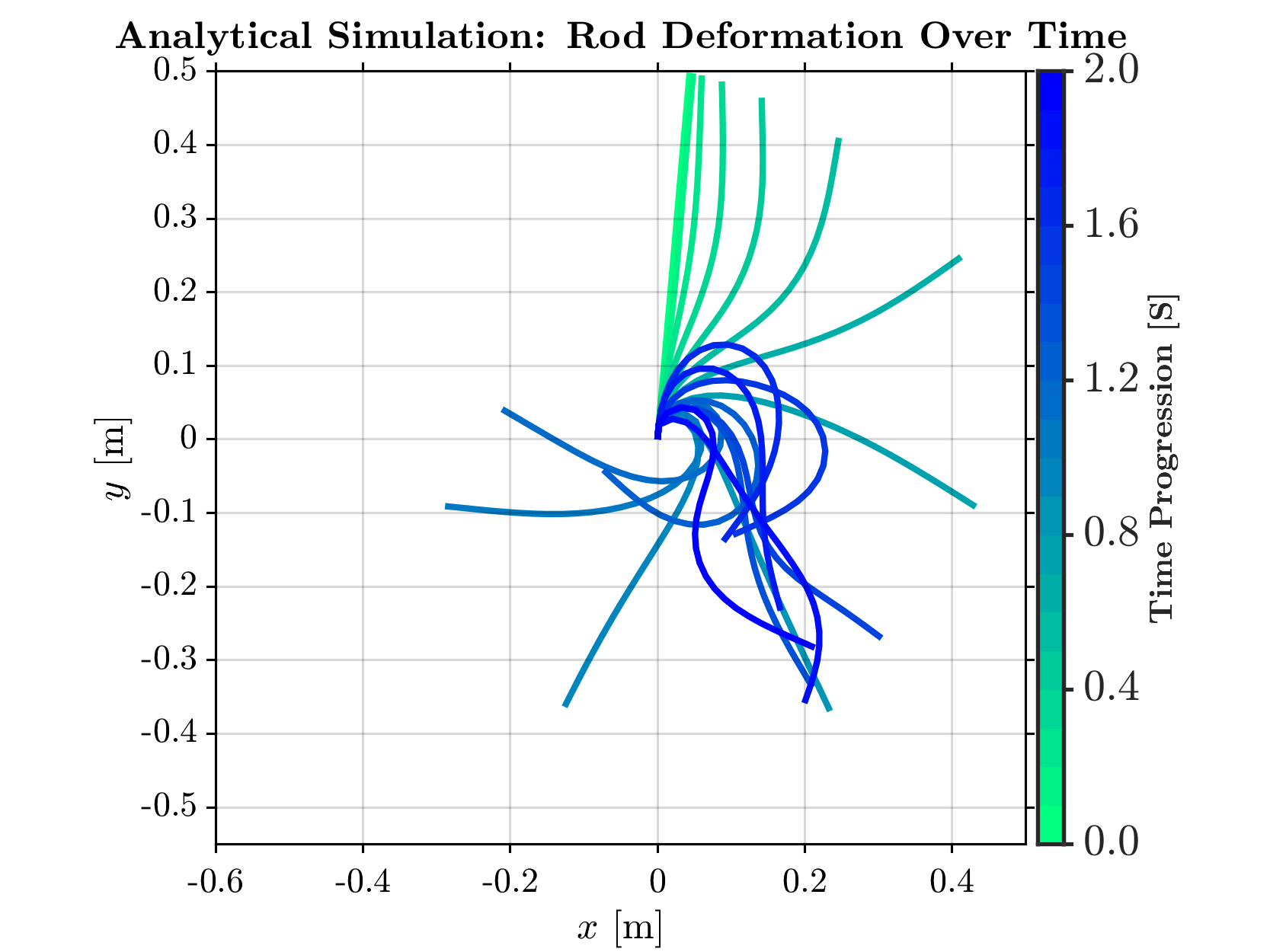}
    \end{minipage}
    \hfill
    \raisebox{-1ex}
{\begin{minipage}[t]{0.48\textwidth}
        
        \includegraphics[width=1.1\textwidth]{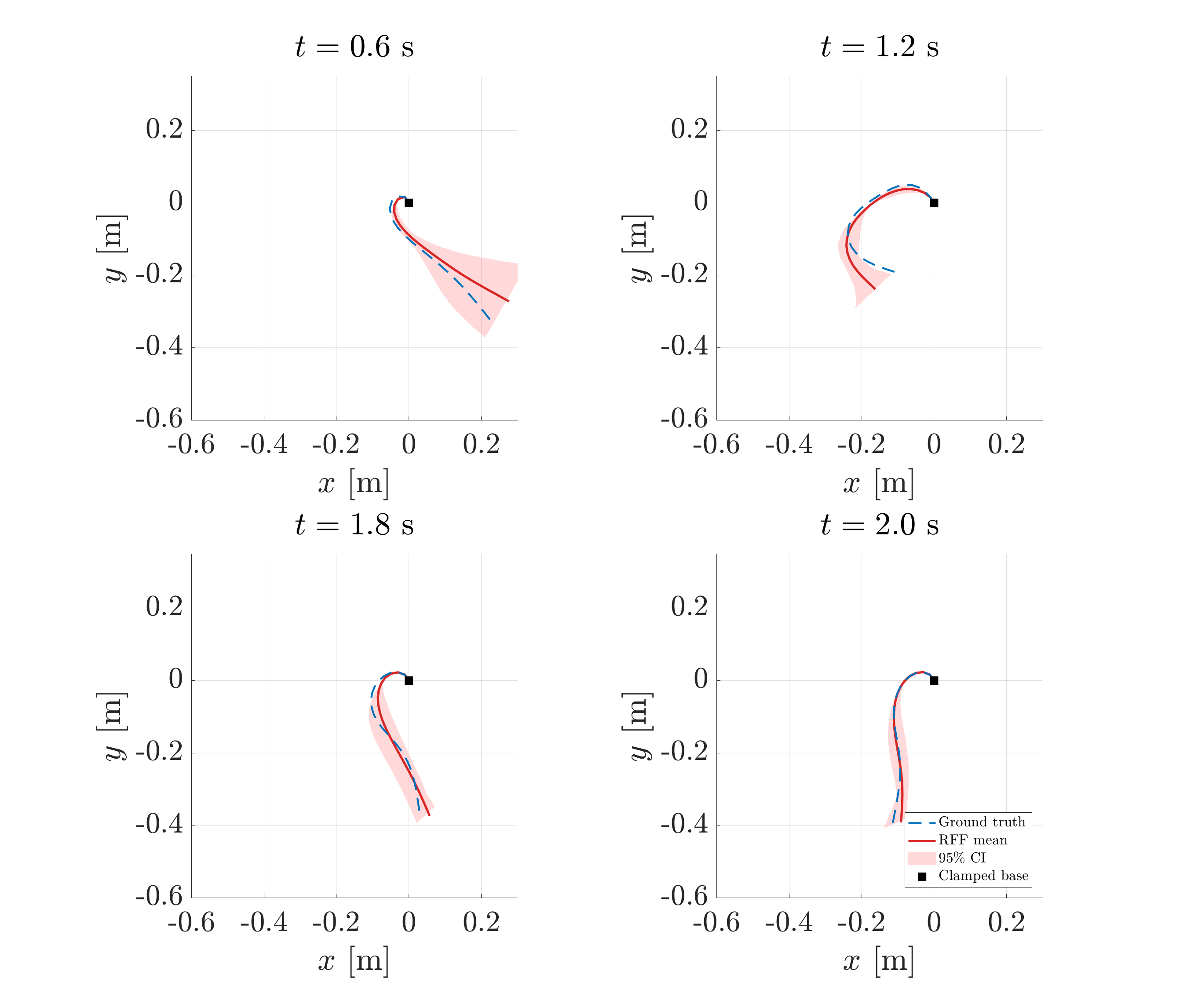}
    \end{minipage}}
    \caption{Left: Ground truth rod motion used for training ($\theta_0 = 85^\circ$). Right:
        CR-GPR prediction vs.\ ground truth at unseen test angle $\theta_0 = 125^\circ$ with $95\%$ CI (shaded). The ground truth (dashed blue) largely lies within the predicted uncertainty band, showing spatially resolved UQ.
    }\vspace{-0.5cm}
    \label{fig:combined_shapes}
\end{figure*}

Fig.~\ref{fig:overlays_all} compares the ground truth and CR-GPR trajectories at representative nodes (base node 1, mid node 13, and tip node 26).  For all state variables, the learned CR-GPR model accurately extrapolates the temporal evolution and phase behavior in comparison with the ground truth simulations.
Fig.~\ref{fig:combined_shapes} (left) shows the ground truth rod deformation over time ($\theta_0 = 85^\circ$), which constitutes the training data set; (right) compares rod configurations at an unseen test angle with 95\% CI bands.
The close match between ground truth (blue) and learned (red) motion across all frames confirms that the CR-GPR model successfully captures spatio–temporal deformation.

\begin{figure}[t]
  \centering
    \includegraphics[width=1\columnwidth]{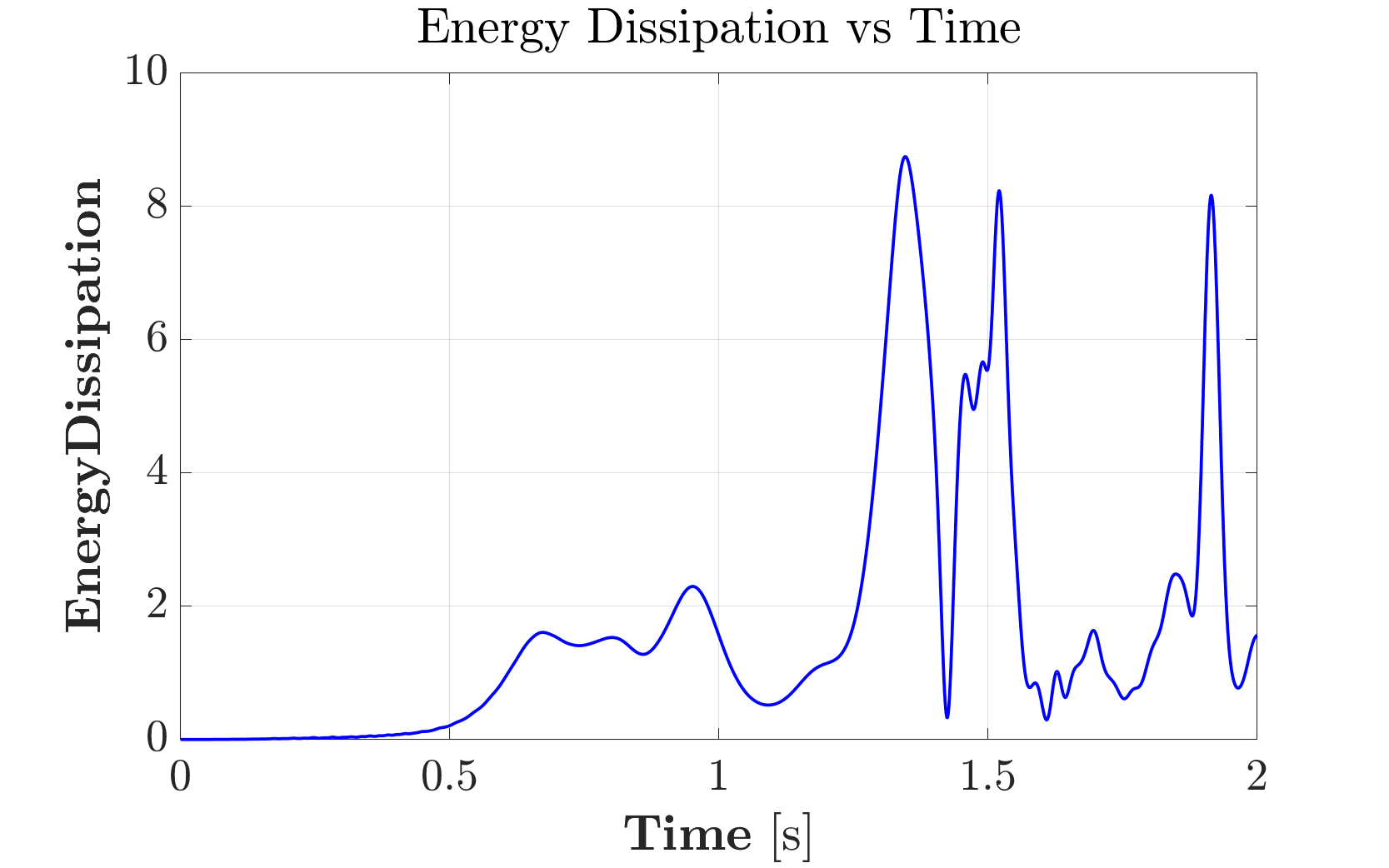}
  \caption{The non-negative energy dissipation confirms the energetic consistency of the CR-GPR.\vspace{-0.5cm}}
  \label{fig:energy_dissipation}
\end{figure}

To verify energetic consistency, we compute the instantaneous dissipated energy $E(t) = (\delta H / \delta z)^{\mathsf{T}} R \, \delta H / \delta z$, where $R \succeq 0$ is the damping matrix. As shown in Fig.~\ref{fig:energy_dissipation}, $E(t)$ is always positive, confirming that the learned model preserves the energy-dissipating structure of the original distributed port--Hamiltonian dynamics.

\begin{figure}[t]
  \centering
  \includegraphics[width=1.05\columnwidth]{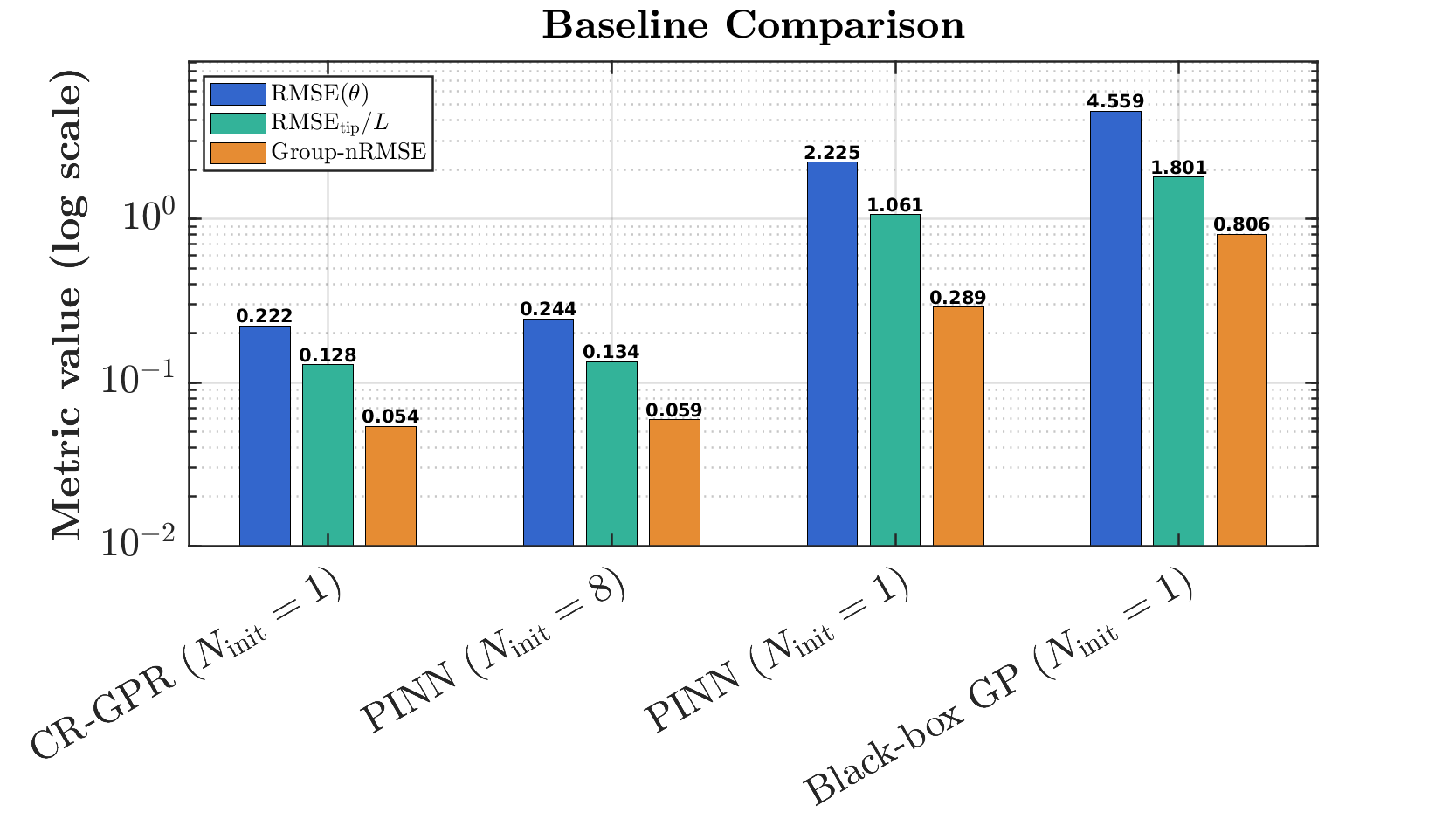}
  \caption{Baseline comparison across six unseen test angles (log scale). Three metrics are shown per method: RMSE($\theta$), RMSE$_{\text{tip}}/L$, and Group-nRMSE. CR-GPR ($N_{\text{init}}=1$) outperforms all baselines across all metrics, including a PINN trained on $8\times$ more data ($N_{\text{init}}=8$).}
  \label{fig:baseline_bar}
\end{figure}

\paragraph{Discussion}
Across all six test angles, CR-GPR achieves a mean $\text{RMSE}(\theta) = 0.222$\,rad and $R^2 = 0.919$ (Table~\ref{tab:multiangle_accuracy}), demonstrating strong generalization from a single training trajectory. The always-positive energy dissipation (Fig.~\ref{fig:energy_dissipation}) confirms passivity, ensuring long-term stability without artificial energy injection. The close match in rod configurations (Fig.~\ref{fig:combined_shapes}) further shows that the learned variational derivatives $\delta H / \delta z$ capture spatial coupling among bending, shear, and translational effects, without explicit knowledge of the governing PDE.
The CR-GPR pipeline assumes only the PHS \emph{structure} ($J$ and $R$ matrices), while the Hamiltonian gradients encoding constitutive behavior are learned entirely from data using generic kernel priors. The strong generalization to six unseen initial conditions (Fig.~\ref{fig:baseline_bar}) confirms that the learned model captures the underlying physics, not merely the training trajectory.


\paragraph{Baseline Comparisons}
To assess the contribution of the port--Hamiltonian structure, we compare CR-GPR against three baselines across the same six unseen initial conditions ($\theta_0 \in \{60^\circ, 125^\circ, 160^\circ, 180^\circ, 225^\circ, 325^\circ\}$), none of which overlap with the training angle ($85^\circ$). All methods are evaluated using the segment orientation error $\text{RMSE}(\theta) = \sqrt{\text{mean}(\epsilon_\theta^2)}$, computed globally over all $N=25$ spatial nodes and all time steps. We also report the tip-position error $\text{RMSE}_{\text{tip}}/L = \sqrt{\text{mean}(\|\mathbf{e}_{\text{tip}}\|^2)}\,/\,L_0$, where $\mathbf{e}_{\text{tip}}$ is the Euclidean tip-position error, and the group-normalized RMSE (Group-nRMSE), defined as the average of $\sqrt{\text{mean}(\epsilon_g^2)}\,/\,\text{range}(z_g)$ across the six state groups $(x, y, \theta, p_x, p_y, p_\theta)$.

\textit{Physics-only PINN.}
We first attempted a purely physics-informed neural network trained solely on the Cosserat rod PDEs~\eqref{eq:ph-compact} without any measurement data. The PINN was unable to reproduce the rod's dynamic behavior: the predicted rod remained nearly stationary regardless of the initial condition. This is consistent with known difficulties of PINNs applied to complex, coupled PDE systems involving large deformations~\cite{wang2024pinnray,sun2022,liu2023pinn}. The nonlinear coupling between spatial derivatives, trigonometric terms, and distributed forces in the Cosserat equations makes the PDE residual loss landscape particularly challenging for gradient-based PINN training. We therefore equip the PINN baselines below with measurement data to provide a fair comparison.

\textit{Baselines with data.}
We consider three data-driven baselines, where $N_{\text{init}}$ denotes the number of initial conditions used for training. The first two baselines train on the same single trajectory ($\theta_0 = 85^\circ$) as CR-GPR, while the third uses $8\times$ more data: (i)~a \emph{black-box GP} ($N_{\text{init}}=1$) that replaces all six Hamiltonian gradient components with independent GPs without the pH structure; (ii)~a \emph{data-assisted PINN} ($N_{\text{init}}=1$); and (iii)~a \emph{data-assisted PINN} ($N_{\text{init}}=8$) trained on eight trajectories spanning $[30^\circ, 330^\circ]$, including $\theta_0 = 85^\circ$. Fig.~\ref{fig:baseline_bar} summarizes the results.

The black-box GP, which lacks the pH structure, achieves an RMSE($\theta$) of $4.559$~rad, over $20\times$ worse than CR-GPR, demonstrating that the port--Hamiltonian inductive bias is essential for generalization. The PINN with $N_{\text{init}}=1$, despite receiving identical training data, yields an RMSE($\theta$) of $2.225$~rad ($10\times$ worse), as the network fails to extrapolate to distant initial conditions. Even the PINN with $N_{\text{init}}=8$, trained on $8\times$ more data, achieves only $0.244$~rad, slightly worse than CR-GPR's $0.222$~rad. This confirms that the pH structure provides a stronger inductive bias than additional data alone, enabling CR-GPR to generalize from a single trajectory while maintaining energy consistency.

\paragraph{Uncertainty Quantification Results}
We draw $S=10$ joint function samples from the GP posteriors via RFF and propagate each through the Cosserat rod ODE ($\Delta t = 10^{-4}$~s) for an unseen test angle ($\theta_0 = 125^\circ$).
Fig.~\ref{fig:combined_shapes}~(right) shows the resulting $95\%$ confidence band (2.5th--97.5th percentile envelope) around the predicted rod configuration at four time instants. The ground truth largely lies within the predicted CI, confirming spatially resolved, physically meaningful uncertainty estimates.

\paragraph{Computational Cost}
Table~\ref{tab:timing} reports wall-clock times (MATLAB R2024b, 16-core AMD workstation, 128\,GB RAM). The total offline training cost (Stages~1--2) is approximately \SI{11}{\minute}, and a single forward simulation takes \SI{2.4}{\minute}. Note that these times could be decreased by using GP approximation techniques and more advanced ODE solvers.

\begin{table}[t]
\caption{Computational cost breakdown of the CR-GPR pipeline.}
\label{tab:timing}
\centering
\setlength{\tabcolsep}{4pt}
\renewcommand{\arraystretch}{1.3}
\begin{tabular}{llr}
\toprule
\textbf{Stage} & \textbf{Description} & \textbf{Time} \\
\midrule
1. GP prefiltering   & Fit state GPs, extract $\dot{z}$          & \SI{3.0}{\minute} \\
2. Gradient learning  & Train 6 $\delta H/\delta z$ models       & \SI{8.0}{\minute} \\
3. Simulation         & Verlet integration ($N_{\text{init}}=1$)  & \SI{2.4}{\minute} \\
4. UQ / RFF sampling  & 10 trajectory samples                     & \SI{36}{\minute}  \\
\bottomrule
\end{tabular}
\end{table}

\paragraph{Ablation: Two-Step vs.\ Single-Step Pipeline}
We replace Stage~1 (GP prefiltering) with central finite differences ($\dot{z}_k = (z_{k+1} - z_{k-1})/2\Delta t$) while keeping Stage~2 and the Verlet simulation unchanged.
Table~\ref{tab:ablation} shows that both pipelines achieve comparable accuracy on clean data, confirming that the pH inductive bias, not the derivative estimation method, is the primary driver of accuracy.
The slightly higher $R^2$ of the single-step pipeline on clean data is expected: finite differences use all training samples at full temporal resolution, whereas GP prefiltering subsamples the data and introduces a small smoothing bias through the kernel prior.
The GP prefiltering stage serves as a \emph{robustness mechanism}: with noisy experimental data, finite differences would amplify sensor noise, whereas GP smoothing suppresses it.

\begin{table}[t]
\caption{Ablation: GP-smoothed (two-step) vs.\ finite-difference (single-step) pipeline. Metrics averaged over six test angles.}
\label{tab:ablation}
\centering
\setlength{\tabcolsep}{4pt}
\renewcommand{\arraystretch}{1.3}
\begin{tabular}{lcccc}
\toprule
\textbf{Pipeline} & \textbf{RMSE($\theta$)} & \textbf{Tip/L} & \textbf{G-nRMSE} & $\mathbf{R^2}$ \\
\midrule
Two-step (GP $\dot{z}$) & 0.2218 & 0.1278 & 0.0540 & 0.9186 \\
Single-step (FD $\dot{z}$) & 0.2107 & 0.1248 & 0.0530 & 0.9840 \\
\bottomrule
\end{tabular}
\end{table}


\section{Conclusion}

This work presented CR-GPR, a framework that combines Cosserat rod theory with Gaussian Process Regression to learn soft robot dynamics while preserving the port--Hamiltonian energy structure. Trained on a single trajectory, CR-GPR generalizes to unseen initial conditions, maintains physical consistency through a non-negative decaying energy profile, and outperforms black-box GP ($20\times$), single-trajectory PINN ($10\times$), and even a PINN with $8\times$ more training data. RFF-based uncertainty quantification further provides spatially resolved confidence intervals that contain the ground truth trajectory. Future work will extend the framework to three-dimensional rod models and leverage the rich structure for passivity-based control design.








\end{document}